%% file: main.tex
\documentclass[10pt,twocolumn,letterpaper]{article}

\usepackage[pagenumbers]{iccv} 
\usepackage{xspace}
\usepackage{cuted}


\definecolor{iccvblue}{rgb}{0.21,0.49,0.74}
\usepackage[pagebackref,breaklinks,colorlinks,allcolors=iccvblue]{hyperref}

\usepackage{bbding}
\usepackage{pifont}
\usepackage{xcolor}
\usepackage{multirow}
\usepackage[utf8]{inputenc}

\definecolor{goldttt}{rgb}{0.1, 0.7, 0.1}  
\definecolor{goldgreen}{rgb}{0.0, 0.5, 0.0}  
\definecolor{goldorange}{rgb}{0.1, 0.7, 0.1}  
\definecolor{graycolor}{rgb}{0.6, 0.6, 0.6}  
\definecolor{suppcolor}{rgb}{0.9, 0.4, 0.4}

\def\paperID{592} 
\def\confName{ICCV}
\def\confYear{2025}

\newcommand{\camerapapername}{I2VControl-Camera\xspace}
\newcommand{\papername}{I2VControl\xspace}

\title{I2VControl: Disentangled and Unified Video Motion Synthesis Control}

\author{
    Wanquan Feng\textsuperscript{1$\dagger$} \quad
    Tianhao Qi\textsuperscript{1,2} \quad
    Jiawei Liu\textsuperscript{1} \quad
    Mingzhen Sun\textsuperscript{1,3} \quad
    Pengqi Tu\textsuperscript{1} \\[1pt]
    Tianxiang Ma\textsuperscript{1} \quad
    Fei Dai\textsuperscript{1} \quad
    Songtao Zhao\textsuperscript{1} \quad
    Siyu Zhou\textsuperscript{1} \quad
    Qian He\textsuperscript{1} \\[1pt] 
    \normalsize{\textsuperscript{1}Intelligent Creation Team, ByteDance \quad \textsuperscript{2}University of Science and Technology of China (USTC)} \\[1pt] 
    \normalsize{\textsuperscript{3}Institute of Automation, Chinese Academy of Sciences (CASIA)}
}

\begin{document}
\maketitle

\begin{strip}
    \vspace*{-18mm}
    \centering
    \includegraphics[width=1\textwidth]{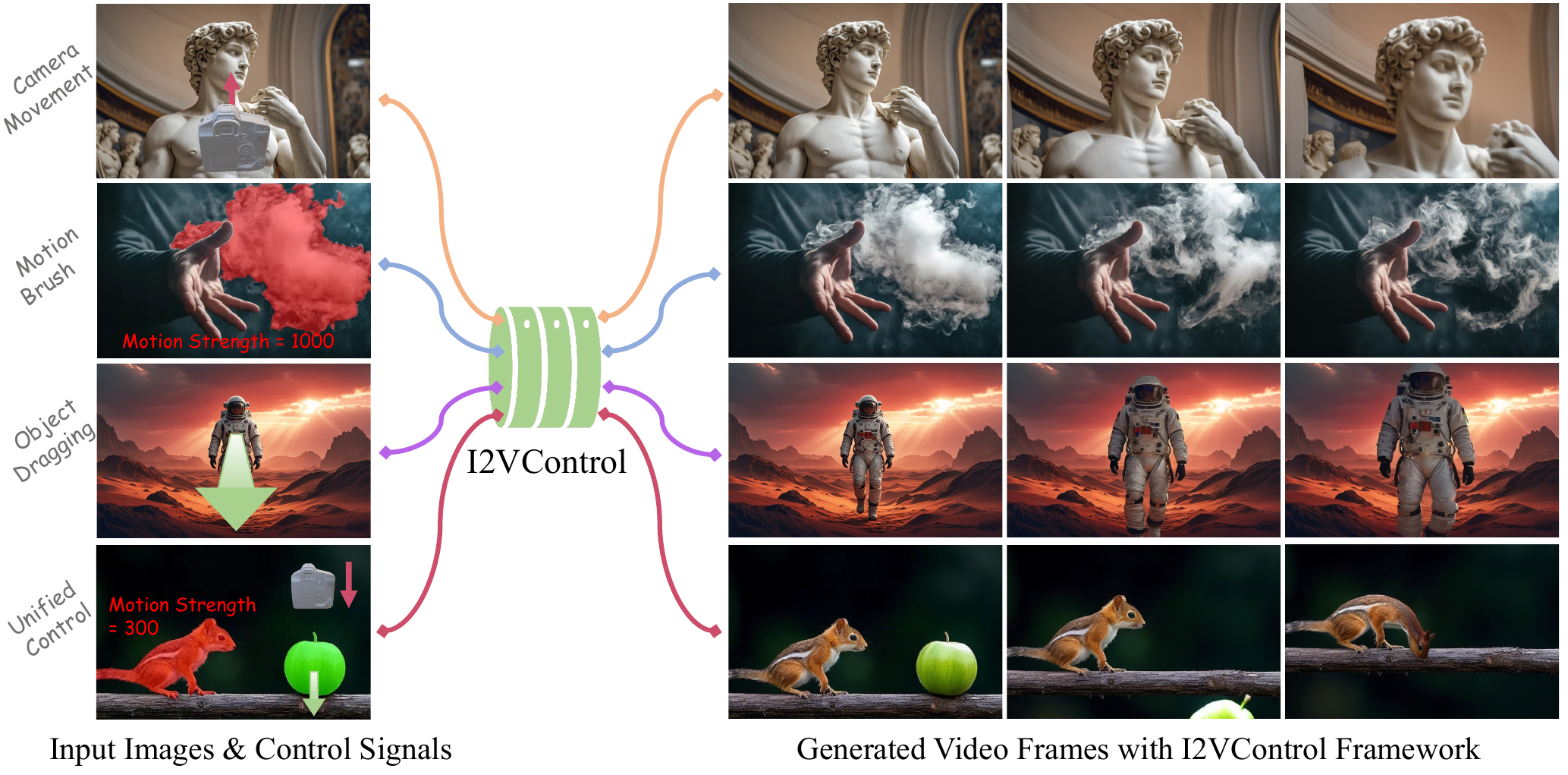}
    \captionof{figure}{We propose \textbf{an all-in-one disentangled and unified framework for image-to-video motion synthesis control}, named \textbf{\papername}. In the illustration, we show several scenarios of controls, including camera movement (camera dollies in and gets closer to the sculpture), motion brush (smoke flows in the wind, with a given motion strength value), object movement (the astronaut walks forward),  and the combination of all above control types (camera tilts down; drag the apple; brush the squirrel). \textbf{Users can select the control modes according to their requirements, where the control modes can be combined without conflict}.}
    \label{teaser}
    \vspace*{-2mm}
\end{strip}

\vspace*{-6mm}
\begin{abstract}
Motion controllability is crucial in video synthesis. However, most previous methods are limited to single control types, and combining them often results in logical conflicts.
In this paper, we propose a disentangled and unified framework, namely \textbf{I2VControl}, to overcome the logical conflicts.
We rethink camera control, object dragging, and motion brush, reformulating all tasks into a consistent representation based on point trajectories, each managed by a dedicated formulation. 
Accordingly, we propose a spatial partitioning strategy, where each unit is assigned to a concomitant control category, enabling diverse control types to be dynamically orchestrated within a single synthesis pipeline without conflicts.
Furthermore, we design an adapter structure that functions as a plug-in for pre-trained models and is agnostic to specific model architectures.
We conduct extensive experiments, achieving excellent performance on various control tasks, and our method further facilitates user-driven creative combinations, enhancing innovation and creativity.
Project page: \href{https://wanquanf.github.io/I2VControl}{https://wanquanf.github.io/I2VControl}.
\end{abstract}

\input{01_introduction}
\input{02_related_work}
\input{03_preliminary}
\input{04_method}

\input{05_experiments}
\input{06_conclusion}

{
    \small
    \bibliographystyle{ieeenat_fullname}
    \bibliography{main}
}


\end{document}

%% file: 01_introduction.tex
\vspace{-2mm}
\section{Introduction}
\label{sec:introduction}

\begin{quote}
\textit{``The whole is greater than the sum of its parts.''}

\hfill -- Aristotle, \textit{Metaphysics}
\vspace{-3mm}
\end{quote}

Motion controllability plays a pivotal role in video synthesis technology~\cite{videoworldsimulators2024,kling,bao2024vidu,wan2.1}, particularly when video creators require meticulously designed spatiotemporal dynamics.
To align generated video motion with human intent — a goal often unmet by text prompts~\cite{clip} due to their limited expressiveness — some approaches have attempted to guide the video synthesis process by injecting additional control signals, such as motion dragging~\cite{wu2024draganything,boximator,jain2024peekaboo}, camera pose~\cite{wang2023motionctrl,he2024cameractrl} and motion brush~\cite{shi2024motion,gen2}. 

\begin{figure}[t]
\centering
\vspace{-8mm}
\includegraphics[width=0.48\textwidth]{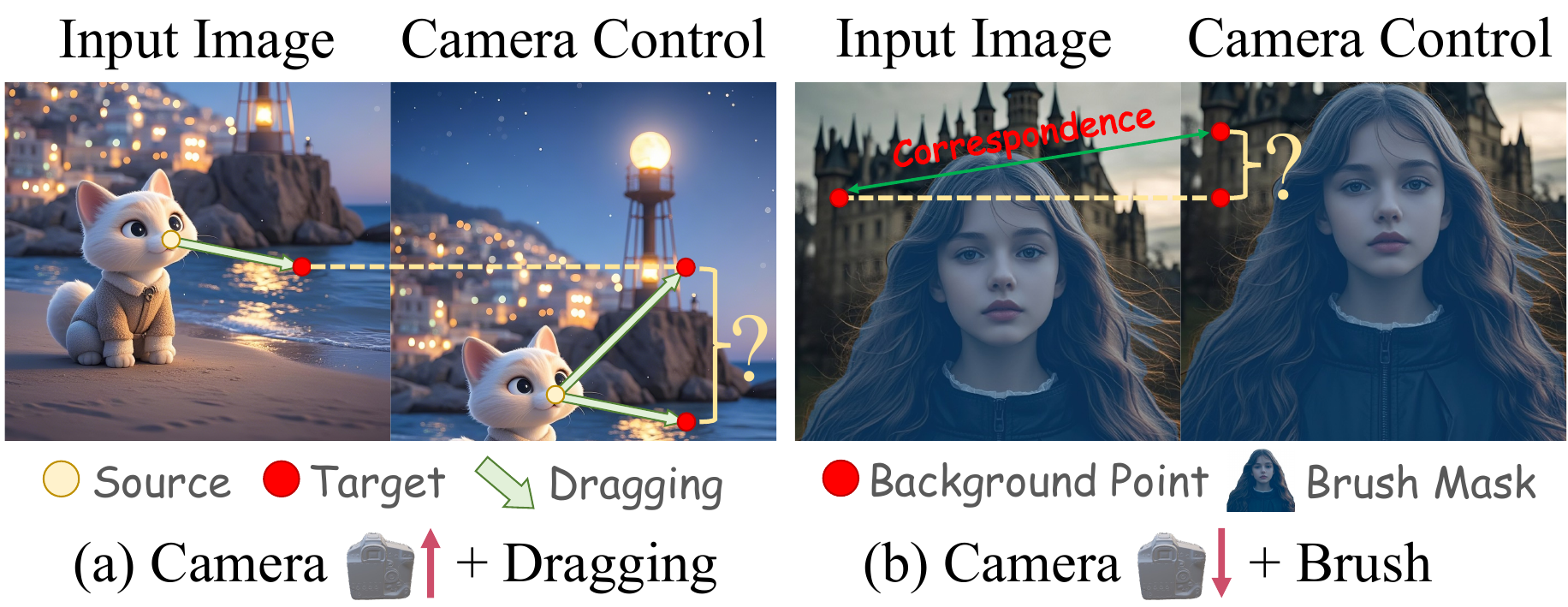}
\caption{Examples of control conflicts in multiple control tasks. In (a), we try to drag the cat and apply camera tilt-up at the same time. We drag the cat towards the ``right + down" direction; however, in the tilt-up view, the target position become much higher than the cat, which is an obfuscation. In (b), we try to employ motion brush and camera tilt-down together. We brush the human region, keeping the background fixed; however, in the tilt-down view, every background pixel moves, which is also a conflict.}
\vspace{-6mm}
\label{fig:conflict}
\end{figure}

Previous motion control approaches often focus on a single type of motion pattern and necessitate the construction of specific data formats (e.g. 3D static dataset RealEstate10K~\cite{zhou2018stereo} for camera control) tailored for certain scenarios, lacking a comprehensive framework for the integration of multiple control types. Simply integrating them into a single inference pipeline would lead to logical conflicts. For instance in Fig.~\ref{fig:conflict} (a), the dragging control requires source and target positions for objects; however, if camera control is added, the target position may differ across different camera views, which could lead to inconsistencies of user intent. Another example in Fig.~\ref{fig:conflict} (b) involves using a motion brush (animate a selected area while keeping other areas fixed); if camera control is added, background pixels would move, creating logical conflict.

This limitation is a significant barrier for complex video synthesis tasks. Consequently, previous single-typed methods, while innovative, do not fully meet the expectations for a satisfactory and intuitive user experience. They often lack the versatility and flexibility required to seamlessly control the video synthesis process, thus providing users with tools that may not be sufficiently adaptable to the wide range of creative demands encountered in video production.

In this paper, we propose a novel video control framework, \textbf{\papername}, designed to integrate multiple control signals within a single cohesive system. To achieve this, we progressively implement three important designs:

\noindent \textbf{$\bullet$ Consistent Representation.} Camera control methods typically use the extrinsic matrix~\cite{wang2023motionctrl,he2024cameractrl} as signal, while dragging utilize sparse points~\cite{wu2024draganything} or bounding boxes~\cite{boximator,jain2024peekaboo} to dictate object movement, and motion brush task often employs optical flow~\cite{shi2024motion}. These representations are entirely disparate, making simultaneous control extremely challenging. Thus, our first step is to unify the representation across these tasks. 
Specifically, we choose dense point trajectories as our unified representation. 
In Sec.~\ref{sec:method}, we will show how we design different formulations of this representation to manage different control types. This unification allows us to integrate different tasks more easily.

\noindent \textbf{$\bullet$ Spatial Partitioning.} A natural observation is that each control intent correlates with a specific spatial region on the input image. The motion brush is applied to a designated area, the object being dragged has its own mask, and camera movement involves a mask formed by all areas not occupied by other controls. During inference, it is easy for users to select these masks, either manually or by using SAM~\cite{kirillov2023segany}. With this mechanism, the framework can support multiple regional controls on one single input image.

\noindent \textbf{$\bullet$ Network Structure.} Despite a consistent representation and partitioning mechanism, a key challenge remains: inputting this information to the network. We need to provide the model with some key information: the partition map, the type for each region, and the control signals for each region. Thanks to the unified representation, each pixel is associated with its point trajectory, which is a regular form that can be easily inputted into the network. Additionally, we convert the partition and control types of each unit into spatial maps, which are concatenated with the trajectory to injected into the network. After several layers of convolution, the input is encoded into tokens whose dimensions is harmonious with the pretrained model. Then we employ some additional trainable attention layers to form an adapter.

Certainly, we would face a significant challenge in effectively simulating user inputs in the training data. Therefore, we devise a data pipeline to align the training videos with our designed control signal, streamlining the training process. Experimental results demonstrate that our framework offers outstanding flexibility and effectiveness across a range of control scenarios. Moreover, it matches or surpasses the top performances of previous methods in one-typed tasks, including dragging, camera control, and motion brush.
In summary, our \textbf{contributions} include:
\begin{itemize}
    \item As far as we know, \textbf{\papername} is the first framework that can conduct camera control, object dragging and motion brush together in one single pipeline without conflicts.
    \item To achieve unified control, we have developed the consistent representation, spatial partitioning mechanism, the adapter network, and a data pipeline for training.
    \item Quantitative and qualitative results are excellent.
\end{itemize}

%% file: 02_related_work.tex
\section{Related Work}
\label{sec:related_work}

\subsection{Text to Video Synthesis}
\label{sec:related_work:t2v}

Early research on text-to-video (T2V) generation utilized VAE~\cite{vaet2v}, GAN~\cite{gant2v}, and sequential generative models~\cite{nuwa,godiva,cogvideo,tats} with limited quality. Recent advancements have been driven by diffusion probabilistic models (DPMs), significantly enhancing video quality~\cite{videoworldsimulators2024,svd}. Initiatives like Imagen Video~\cite{imagenvideo} and Make-A-Video~\cite{makeavideo} achieved initial low-resolution video outputs from text, which were refined through spatial super-resolution and temporal interpolation. To manage computational complexity, video auto-encoders~\cite{videoldm,pvdm,lvdm,walt} encoded videos into compact latent spaces. Notably, Phenaki~\cite{phenaki} developed a variable-length video auto-encoder utilizing causal attention. A hybrid approach~\cite{magicvideo2,emuvideo,vidm} combined text-to-image and image-to-video synthesis to improve the video quality.

\subsection{Image to Video Synthesis}
\label{sec:related_work:i2v}

Image-to-video (I2V) generation converts static images into videos. VideoCrafter1~\cite{chen2023videocrafter1} used a dual cross-attention layer to integrate CLIP embeddings of images in DPM. I2V-Adapter~\cite{guo2024i2v} built on this approach by facilitating interactions between the initial image and noisy frames via cross-frame attention. Animate Anyone~\cite{hu2024animate} addressed the loss of detail in CLIP embeddings by using a network identical to DPMs for image condition injection. DreamVideo~\cite{wei2024dreamvideo} used a ControlNet-like structure for high-fidelity outcomes. Additionally, SEINE~\cite{chen2023seine}, PixelDance~\cite{zeng2024make}, and PIA~\cite{zhang2024pia} expanded DPM input channels by combining image latents with noisy latents to preserve details. DynamicRafter~\cite{xing2023dynamicrafter}, I2VGen-XL~\cite{zhang2023i2vgen}, and SVD~\cite{blattmann2023stable} mixed various image condition injection techniques to handle precise details.

\subsection{Video Synthesis Controllability}
\label{sec:related_work:controllability}

With the development of video generation, some motion control technologies have also been proposed recently. For camera control, MotionCtrl~\cite{wang2023motionctrl} and CamereCtrl~\cite{he2024cameractrl} employed an adapter for the extrinsic matrix of each frame to control the camera movement, and \camerapapername~\cite{i2vcontrolcamera} proposed a motion strength controller for further adjustive motion dynamic. 
Boximator~\cite{boximator} and Direct-A-Video~\cite{yang2024direct} utilized the trajectory of the bounding box of the object to guide the object motion. 
DragNVWA~\cite{dragnuwa}, DragAnything~\cite{wu2024draganything}, MOFA-Video~\cite{niu2024mofa} utilized sparse control points to guide object motion.
Motion brush functionality enables users to drive motion in specific regions via a provided mask, with a scalar motion strength value, which is supported in works like Motion-I2V~\cite{shi2024motion} and Gen-2~\cite{gen2}. We propose to unify multiple types in one single framework.

%% file: 03_preliminary.tex
\section{Preliminary and Rethinking}
\label{sec:preliminary}

\subsection{Dual Task}
\label{sec:dual}

\begin{figure}[t]
\centering
\vspace{-4mm}
\includegraphics[width=0.48\textwidth]{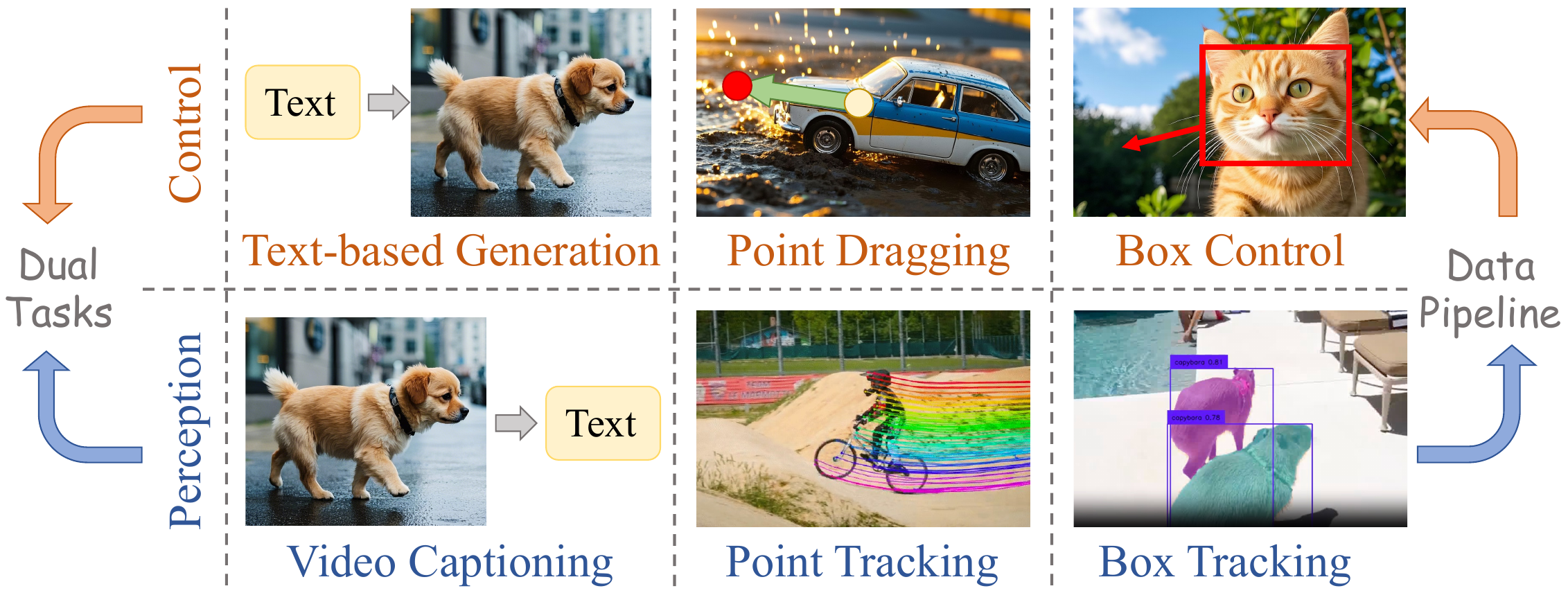}
\vspace{-6mm}
\caption{Examples of dual tasks, where the dual perception task serves as the data pipeline of the controlling generation task.}
\vspace{-6mm}
\label{fig:dual}
\end{figure}

We first analyze two key observations regarding controllable generation, which is inspiring for algorithm design:

\noindent \textbf{$\bullet$ Every controllable generation task has a perception task as its dual task.} We show examples in Fig.~\ref{fig:dual}, including ``text-controlled generation and captioning," ``point dragging and point tracking," and ``box control and box detection." Other instances encompass ``camera pose control and 6-DoF visual odometry," as well as ``motion brush control and motion strength analysis," among others.

\noindent \textbf{$\bullet$ Dual perception task serves as data pipeline for controllable generation task.} For example, we can apply captioning to obtain text-video pairs to train T2V, or track points in training videos to construct training data for dragging framework. For other tasks, the logic is similar. This observation inspires us to approach from a perception perspective, starting with a video to design a pipeline that exactly serves as the dual task to our unified control.

\subsection{Dense Trajectory Representation}
\label{sec:trajectory_function}

As described in Sec.~\ref{sec:introduction}, we choose dense trajectory as our control representation. We conceptualize video and model trajectory function inspired by I2VControl-Camera~\cite{i2vcontrolcamera}. 
A video can be conceptualized as a sequence of frames in time range $\lambda \in [0,T]$, each capturing a projection of a deformable 3D world through a camera with potential camera movement. We denote the domain of 3D points captured by the first frame (at time moment $\lambda=0$) as $\Omega \subseteq \mathbb{R}^3$.

\noindent \textbf{$\bullet$ Trajectory Function in World System.} We use the camera coordinate system of the first frame as the world coordinate system. So we can define a trajectory function in the world coordinate system, $\mathcal{D}(\lambda, \mathbf{x}): [0, T] \times \Omega \to \mathbb{R}^3$, where $\mathcal{D}(\lambda, \mathbf{x})$ represents the position of the point $\mathbf{x} \in \Omega$ at time $\lambda$. At the initial moment ($\lambda=0$), we have $\mathcal{D}(0, \mathbf{x}) = \mathbf{x}$. 

\noindent \textbf{$\bullet$ Trajectory Function in Camera System.} Further, we convert the trajectory function into the camera coordinate system. At each time $\lambda$, we denote the camera extrinsic as $\mathcal{E}_\lambda \in SE(3)$. The transformation from the world coordinate system (i.e., the coordinate system of the first frame) to the camera coordinate system at time $\lambda$ is given by the inverse of $\mathcal{E}_\lambda$, denoted as $\mathcal{E}^{-1}_\lambda$.
Thus, the transformation of the trajectory function $\mathcal{D}$ from the world coordinate system to the camera coordinate system can be expressed as:
\begin{align}
\mathcal{F}(\lambda, \mathbf{x}) = \mathcal{E}^{-1}_\lambda \circ \mathcal{D}(\lambda, \mathbf{x}),
\end{align}
where $\mathcal{F}(\lambda, \mathbf{x})$ represents the position of $\mathbf{x}$ in camera system at time $\lambda$. 
Considering $\mathcal{E}_0=\mathcal{I}$, we can deduce that:
\begin{align}
\mathcal{F}(0, \mathbf{x}) = \mathcal{E}_0^{-1} \circ \mathcal{D}(0, \mathbf{x}) = \mathcal{I} \circ \mathcal{D}(0, \mathbf{x}) = \mathbf{x},
\end{align}
which proves that $\mathcal{F}(\lambda, \mathbf{x})$ is still a trajectory function representing the point motions from their initial positions.

\noindent \textbf{$\bullet$ Motion Strength.} We refer to \camerapapername~\cite{i2vcontrolcamera} and define a general motion strength function. Consider an arbitrary domain $\Gamma \subset \mathbb{R}^3$ and trajectory function $\mathcal{H}(\lambda, \mathbf{x}): [0, T] \times \Gamma \to \mathbb{R}^3$, we define the motion strength $\mathcal{M}$ as:
\begin{align}
\mathcal{M}(\Gamma, \mathcal{H}, \lambda) &= \frac{1}{|\Gamma|} \int_{\Gamma} \| \frac{\partial \mathcal{H}(\mathbf{x},\lambda)}{\partial \lambda} \|_2 \, d \mathbf{x},
\label{eq:motion_strength_function}
\end{align}
yielding a scalar that denotes the average speed of points.

%% file: 04_method.tex
\section{Method}
\label{sec:method}

In this section, we present our method details. We introduce our video representation and partitioning in Sec.~\ref{sec:method:video_representation}, the control signal construction in Sec.~\ref{sec:control_signal_construction}, our dual task pair (data pipeline; controllable inference) in Sec.~\ref{sec:our_dual}, our network and training process in Sec.~\ref{sec:method:model_structure}. \textit{Note: In this section, we continue to use the notation established in Sec~\ref{sec:trajectory_function}}.

\subsection{Video Representation and Partitioning}
\label{sec:method:video_representation}


We first introduce the motion unit partitioning, and then define the unit-wise trajectory function formulation.

\noindent \textbf{$\bullet$ Motion Units.} 
Typically, the domain captured by the camera is composed of multiple independent motion parts instead of one total part, which encourages us to divide the entire domain $\Omega$ into multiple independent ``motion units'':
\begin{align}
\Omega = \bigsqcup\limits_{p=0}^{P} \Omega^{(p)} = \bigsqcup\limits_{p=0}^{P} (\chi^{(p)} \odot \Omega),
\end{align}
where $p \in \{0,1,...,P\}$ denotes the unit index, $\Omega^{(p)}$ denotes the $p$-th unit, $\chi^{(p)}$ denotes the mask of the $p$-th unit. Based on the properties and control situations, we summarize these units into three categories, defined as: \textbf{brush-units}, \textbf{drag-units}, and the \textbf{borderland}. Both brush-units and drag-units belong to the foreground, and the difference between them lies in the control type. The brush-units are controlled by scalar value motion strength only, while the drag-units can be controlled by $6$-DOF ($6$ degrees of freedom) motion guidance (a rigid transformation within the group $SE(3)$) and additional motion strength. During training, we randomly set a foreground part as one of the two categories. During inference, the choice is made by the users. The region not covered by the brush-units and the drag-units is summarized as a single part called borderland. During the training and inference process, we always consider the borderland part as the first part with index $0$, both the brush-units and the drag-units are assigned indices greater than $0$.

\noindent \textbf{$\bullet$ Unit-wise Trajectory Function.} 
To handle the unit-wise motion, on each unit $\Omega^{(p)}$, we decompose the trajectory function into two terms: a rigid term and an additional term:
\begin{align}
\mathcal{D}(\lambda, \mathbf{x}) &= \mathcal{R}^{(p)}_{\lambda}\circ \mathbf{x} + \mathcal{G}^{(p)}(\lambda, \mathbf{x}), \; \forall \mathbf{x} \in \Omega^{(p)},  p \in \{0...,P\},
\label{eq:decomposition}
\end{align}
where $\mathcal{R}^{(p)}_{\lambda} \in SE(3)$ is a rigid transformation of the initial point set $\Omega^{(p)}$, and $\mathcal{G}^{(p)}(\lambda, \mathbf{x})$ is the additional term showing the difference between $\mathcal{R}^{(p)}_{\lambda}$ and $\mathcal{D}(\lambda, \mathbf{x})$. 

\subsection{Control Signal Construction} 
\label{sec:control_signal_construction}
In this section, we describe how we construct the control signals.
Each unit is assigned with a $\mathcal{R}^{(p)}_{\lambda}$ and motion strength $m_{\lambda}^{(p)}$. Now we introduce how we construct them:

\noindent $\bullet$ On the \textbf{borderland}, we simply set:
\begin{align}
\mathcal{R}^{(p)}_{\lambda} \equiv \mathcal{I}, \;
m_{\lambda}^{(p)} \equiv 0.
\end{align}
Note that, despite setting $m_{\lambda}$ to 0 during both training and testing, this does not imply that the borderland is completely stationary. On the contrary, it may still exhibit some motion dynamics, ensuring the borderland integrates naturally and seamlessly with the drag-units and brush-units to create a coherent and lifelike video scene. In our constructed training data, the borderland may contain natural small motions, which are also present in the inference results.

\noindent $\bullet$ On the \textbf{brush-units}, we set: 
\begin{align}
\mathcal{R}^{(p)}_{\lambda} \equiv \mathcal{I}, \;
m_{\lambda}^{(p)} = \mathcal{M}(\Omega^{(p)},  \mathcal{G}^{(p)}, \lambda).
\end{align}
Similar to the borderland, we set $\mathcal{R}^{(p)}_{\lambda} \equiv \mathcal{I}$ on brush-units. Instead, users only need to provide the value of $m_{\lambda}^{(p)}$. We compute the motion strength $\mathcal{M}$ with Eq.~\eqref{eq:motion_strength_function}, which indicates the motion deviation from the identity transformation $\mathcal{I}$. During inference, users can designate an object by drawing a mask and make it move by simply classifying the region as a brush-unit and providing a motion strength value.

\noindent $\bullet$ On the \textbf{drag-units}, we set:
\begin{align}
\mathcal{R}^{(p)}_{\lambda} = \underset{\mathcal{R} \in SE(3)}{\arg\min} &\int_{\Omega^{(p)}} \|\mathcal{D}(\lambda, \mathbf{x}) - \mathcal{R} \circ \mathbf{x}\|^2 d \mathbf{x}, \\
m_{\lambda}^{(p)} &= \mathcal{M}(\Omega^{(p)}, \mathcal{G}^{(p)}, \lambda).
\end{align}
We set $\mathcal{R}^{(p)}_{\lambda}$ as the rigid transformation that best ``fits'' the trajectory, which can be obtained by solving a least-squares problem. In other words, $\mathcal{R}^{(p)}_{\lambda}$ serves as a satisfactory ``proxy'' for the motion in this part. Considering its rigid nature, which can be described using just 6 degrees of freedom, it is also user-friendly. The motion strength $m_{\lambda}^{(p)}$ is computed with the same formulation as the brush-units. However, for the drag-units, $m_{\lambda}^{(p)}$ indicates the motion deviation from the rigid fitting result instead of the identity transformation. During inference, users can control the motion of an object by drawing its mask, designating the region as a drag-unit, and providing both the rigid transformation and the additional motion strength value.

In addition to the unit-wise definitions, another important control is the camera control, denoted as $\mathcal{E}_{\lambda}$. This control is shared by all units and is also represented as a 6-DOF signal, ensuring user-friendliness. Finally, we can compute the \textbf{point trajectory} in the camera coordinate system:
\begin{align}
\mathbf{T}_{\lambda} = \bigsqcup\limits_{p=0}^{P}\Pi(\mathcal{E}_{\lambda}^{-1} \circ  \mathcal{R}^{(p)}_{\lambda} \circ (\chi^{(p)} \odot \Omega)), \; \lambda \in [0,T],
\label{eq:trajectory}
\end{align}
where $\Pi(\cdot)$ means the projection operation. Further, we organize the \textbf{motion strength map} as:
\begin{align}
\mathbf{M}_{\lambda} = \bigsqcup\limits_{p=0}^{P} (m_{\lambda}^{(p)} \cdot (\chi^{(p)} \odot \Omega)), \; \lambda \in [0,T],
\label{eq:ms}
\end{align}
To represent the unit index, we construct a \textbf{partition map}:
\begin{align}
\mathbf{P}_{\lambda} = \bigsqcup\limits_{p=0}^{P} (p \cdot (\chi^{(p)} \odot \Omega)), \; \lambda \in [0,T],
\label{eq:indexmap}
\end{align}
Furthermore, it is essential for the network to recognize the category of each unit. We assign $c^{(p)}$ the values 0, 1, and 2 when the $p$-th unit is classified as borderland, drag-part, and brush-part, respectively. So we define a \textbf{category map}:
\begin{align}
\mathbf{C}_{\lambda} = \bigsqcup\limits_{p=0}^{P} (c^{(p)} \cdot (\chi^{(p)} \odot \Omega)), \; \lambda \in [0,T],
\label{eq:categorymap}
\end{align}
Finally, we concatenate the aforementioned tensors to form the ultimate control signal $(\mathbf{T}_{\lambda}, \mathbf{M}_{\lambda}, \mathbf{P}_{\lambda}, \mathbf{C}_{\lambda})$.

\begin{figure}[t]
	\centering
    \vspace{-8mm}
	\includegraphics[width=1\columnwidth]{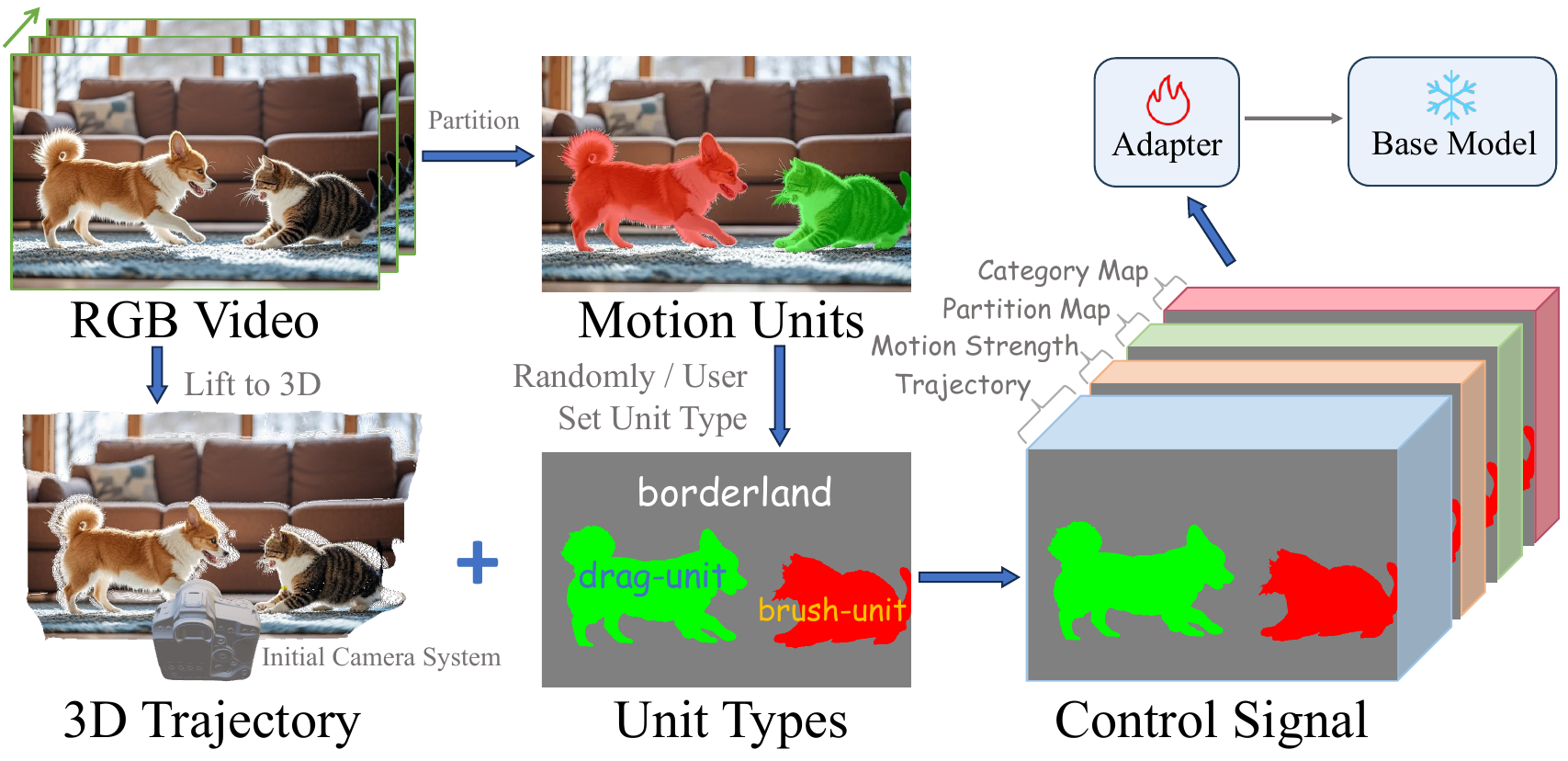}
	\caption{Our data pipeline to convert RGB video into control signals and conduct the training. For more details, please see Sec.~\ref{sec:our_dual}.}
	\label{fig:partition}
    \vspace{-5mm}
\end{figure}

\subsection{Our Dual Task Pair} 
\label{sec:our_dual}

As discussed in Sec.~\ref{sec:dual}, we introduce our dual task pair.

\begin{figure}[t]
\centering
    \vspace{-8mm}
\includegraphics[width=0.48\textwidth]{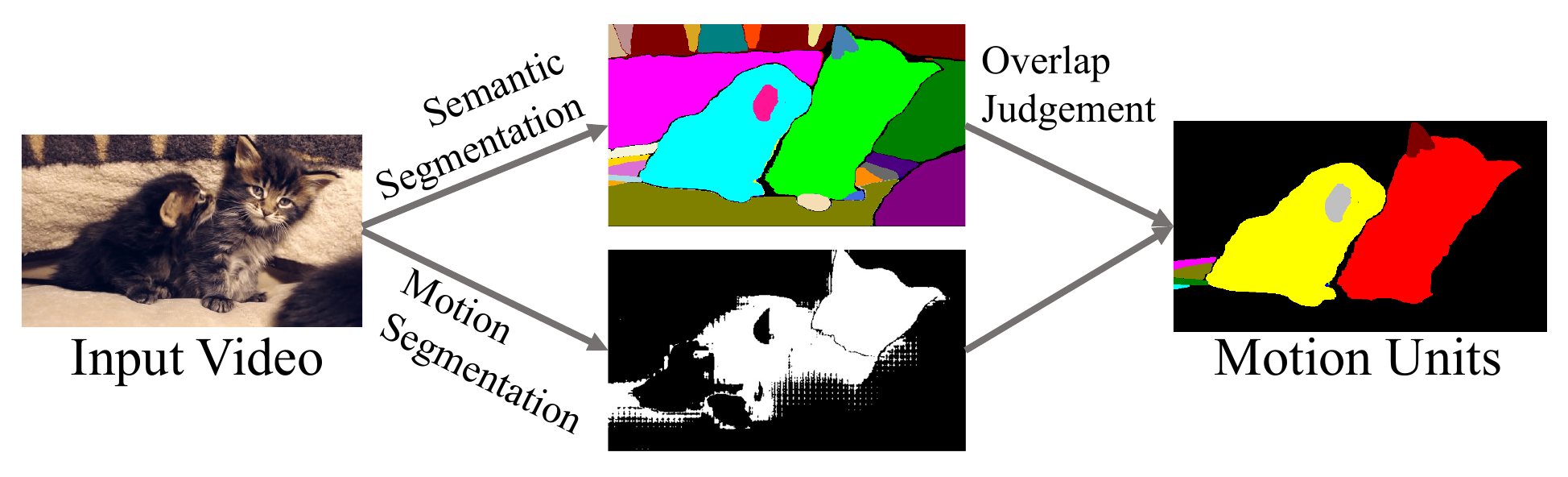}
    \vspace{-8mm}
\caption{The process of obtaining the motion units.}
    \vspace{-5mm}
\label{fig:sam}
\end{figure}

\begin{figure}[h]
\centering
    \vspace{-2mm}
\includegraphics[width=0.47\textwidth]{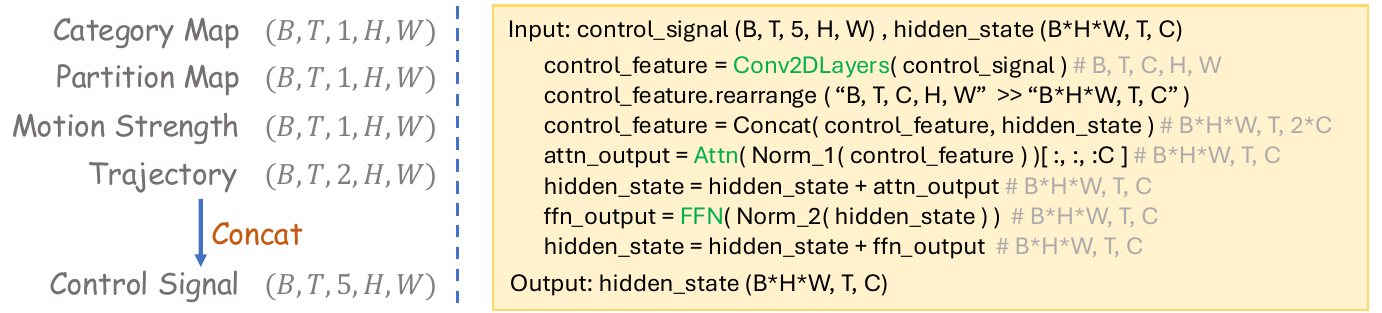}
    \vspace{-2mm}
\caption{Details of adapter architecture and calculation.}
    \vspace{-6mm}
\label{fig:adapter}
\end{figure}

\noindent \textbf{$\bullet$ Data Pipeline.} 
The data pipeline (see Fig.~\ref{fig:partition}) starts with a RGB video. We first conduct the spatial partitioning to obtain the motion units. 
Initially, we use SAM~\cite{kirillov2023segany} to partition the first frame into distinct parts, depicted in the upper branch in Fig.~\ref{fig:sam} for semantic segmentation. We then compute a dynamic mask, as shown in the lower branch in Fig.~\ref{fig:sam} for motion segmentation. Each part extracted from the semantic segmentation map is then evaluated for its intersection with the dynamic mask; areas that overlap more than $50\%$ are selected as brush/drag units, while the remaining areas are designated as borderland.
We employ SpatialTracker~\cite{SpatialTracker} to track $\mathcal{F}(\lambda, \mathbf{x})$, and then follow \camerapapername~\cite{i2vcontrolcamera} to obtain $\{\mathcal{E}_{\lambda}\}_{\lambda \in [0,T]}$ and $\mathcal{D}(\lambda, \mathbf{x}) = \mathcal{E}_{\lambda} \circ \mathcal{F}(\lambda, \mathbf{x})$.
Once we get $\mathcal{D}(\lambda, \mathbf{x})$, we can apply the procedure detailed in Sec.~\ref{sec:control_signal_construction} to compute the control signal tensor. We introduce the details of $(\mathbf{T}_{\lambda}, \mathbf{M}_{\lambda}, \mathbf{P}_{\lambda}, \mathbf{C}_{\lambda})$ and the process of integrating them into the network in Fig.~\ref{fig:adapter}.

\noindent \textbf{$\bullet$ Unified Control: Inference.} During inference, as a user, we need to construct the control signal consistent with the output of data pipeline. We first use the Unidepth~\cite{piccinelli2024unidepth} metric depth estimation method to obtain the initial point set $\Omega$. 
Users then interact with our system to provide further input. 
They can first select the motion units of interest. Specifically, users can employ a bounding box and text description to select a small patch, and then set it as either a brush-unit or a drag-unit. If set as a drag-unit, users are required to provide a 6-DOF input to control the movement of that patch. Users can repeatedly select multiple patches until they have completed their design.
Additionally, users can input an overall camera movement, which does not conflict with the motion of the selected patches. Our system provides a preview interface, allowing users to see the results of the drag and camera movement in real-time, achieving convenient ``\textit{what you see is what you get}'' experience.

\subsection{Network and Training}
\label{sec:method:model_structure}

As mentioned above, we apply an adapter network structure. We show the architecture and pseudo code of adapter in Fig.~\ref{fig:adapter}, where the green layers are trainable. Taking the $(T, 5, H, W)$-shaped tensor as input, we first apply several convolutional layers to convert the input to tokens and then concatenated them with the tokens in the original diffusion process before self-attention computation. After computing self-attention, the additional parts added during concatenation are removed to restore the original token shape. The restored token is added back to the original diffusion token, thus completing the adapter calculation process. 

While training, we extract control signal from ground truth video, then pass it as input to the adapter. Except for inserting the above adapter structure, we keep all other training settings consistent with the original base model, including the loss (only MSE loss) and scheduler.

\begin{figure*}[htb]
	\centering
    \vspace{-8mm}
	\includegraphics[width=2\columnwidth]{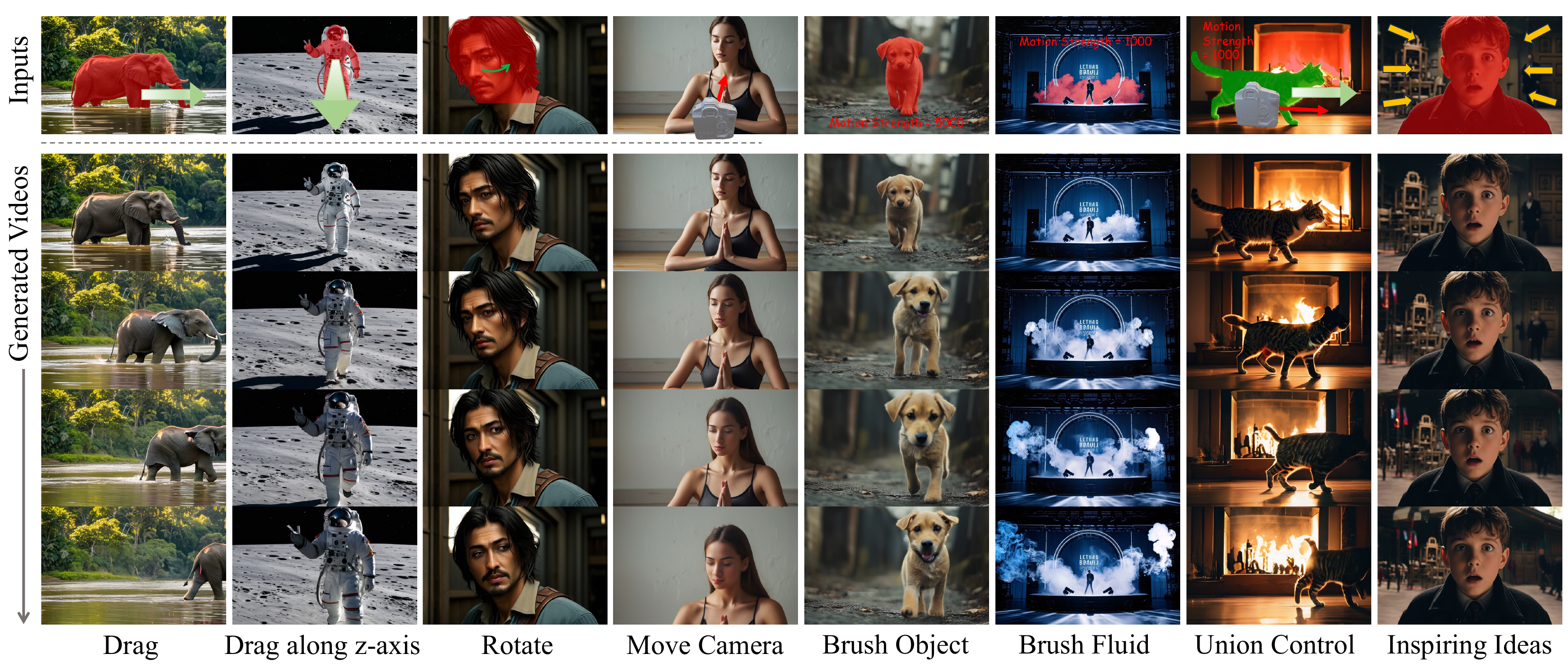}
    \vspace{-3mm}
	\caption{We list some (not all) capabilities of \papername. The first two columns illustrate dragging operations within the xy-plane and along the z-axis, respectively. The third column demonstrates rotational capabilities, and the fourth showcases camera movement control. Columns five and six depict motion brush effects. The seventh column presents a combined control including camera movement, dragging, and motion brush. The eighth column shows a creative user idea, a Hitchcockian camera movement, implemented with \papername.}
    \vspace{-4mm}
	\label{fig:diversity}
\end{figure*}

%% file: 05_experiments.tex
\section{Experiments}
\label{sec:experiments}

\begin{figure*}[htb]
	\centering
	\includegraphics[width=2.0\columnwidth]{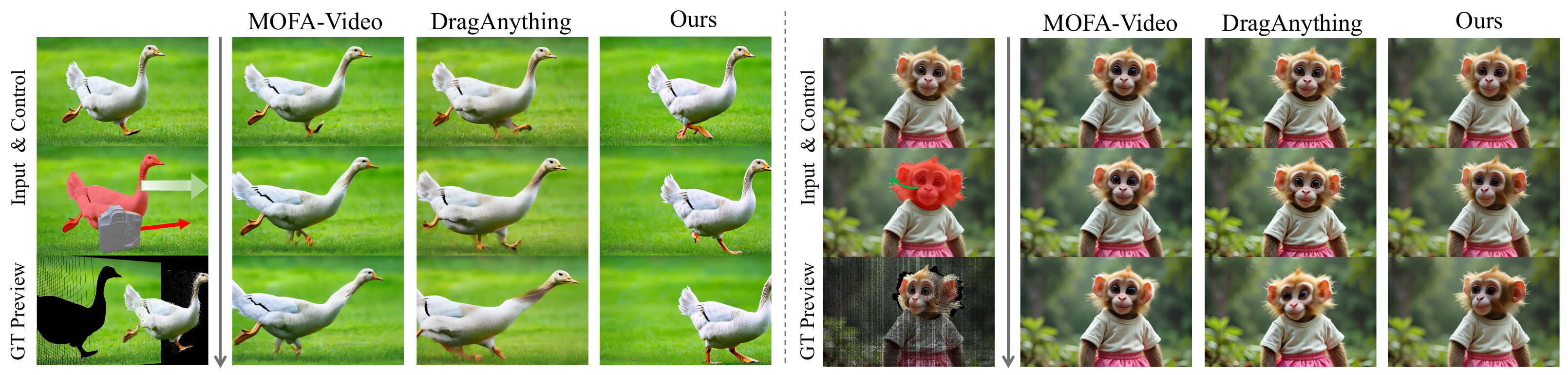}
    \vspace{-2mm}
	\caption{Comparison on dragging control task.}
    \vspace{-2mm}
	\label{fig:trajectory}
\end{figure*}

In this section, we introduce our experiments. Sec.~\ref{sec:experiments:settings} presents our implementation details and settings. Sec.~\ref{sec:experiments:Comprehensiveness} presents the feature comprehensiveness comparison. In Sec.~\ref{sec:experiments:comparisons}, we show our results and experimental comparisons. Please visit our \href{https://github.com/iccv2025sub592/sub592}{anonymous page} for video results.

\subsection{Settings}
\label{sec:experiments:settings}

\noindent \textbf{Implementation Details.} In this work, we utilize an Image-to-Video variant of the Magicvideo-V2~\cite{magicvideov2} as our foundational model ($24$ frames, resolution $704 \times 448$). 

\noindent \textbf{Datasets.} We train on a  training set of $970$K video clips. 
To ensure rigorous comparison, we conduct comprehensive evaluation on $6$ datasets. To maintain consistency with prior research, we adopt $3$ widely-used existing datasets. Additionally, to provide a richer and more convincing comparison, we also curate $3$ new datasets to address certain gaps in the functionalities of previous benchmarks.
We invite several real-world users familiar with video creation to construct high-quality manual control signals including segmentation mask, camera movement and object motion, completely simulating user operation. The testing sets are summarized as:
\begin{itemize}
  \item \textbf{VIPSeg}~\cite{miao2022large}. Following previous works~\cite{wu2024draganything,zhou2024trackgo}, we test on VIPSeg (all testing samples with at least $24$ frames).
  \item \textbf{RealEstate10K}~\cite{zhou2018stereo}. Following previous works~\cite{wang2023motionctrl,he2024cameractrl,i2vcontrolcamera}, we use the testing set of RealEstate10K.
  \item \textbf{MO}~\cite{i2vcontrolcamera}. Following previous work~\cite{i2vcontrolcamera}, we use MO dataset, containing rich camera movement and motions.
  \item \textbf{Manual Dragging}. We invite real-world users to label samples of camera movement and object motion.
  \item \textbf{Object Brush}. We invite users to label motion masks on input images of object motion (e.g., humans, animals). 
  \item \textbf{Fluid Brush}. We invite users to label motion masks on input images of fluid effect samples (e.g., smoke, fire). 
\end{itemize}

\noindent \textbf{Metrics.} For a rigorous comparison, we strictly adopt metrics from previous works. To judge camera control effect, we follow previous works~\cite{wang2023motionctrl,he2024cameractrl,i2vcontrolcamera}, using \texttt{RotErr} and \texttt{TransErr}. To judge dragging, we refer to previous works~\cite{wu2024draganything,zhou2024trackgo}, using \texttt{ObjMC}. To judge motion brush, we firstly refer to previous work~\cite{i2vcontrolcamera} to compute a motion strength score \texttt{MSC}, and then refer to SAM~\cite{kirillov2023segany} to employ a \texttt{IoU} metric to evaluate whether the generated motion follows the user input mask.
Moreover, we additionally employ \texttt{FID} and a \texttt{User} study score to measure the quality.

\begin{table}[t]
	\small
	\setlength\tabcolsep{5pt}
	\begin{center}
		\begin{tabular}{l|ccc}
			\bottomrule
			&  Motion   & Camera  & Motion \\
			&  Drag  & Extrinsics & Brush \\ \hline \hline
	Boximator~\cite{boximator}   & \textcolor{goldorange}{\textbf{\checkmark}}   & \textcolor{graycolor}{\textbf{$\times$}}  & \textcolor{graycolor}{\textbf{$\times$}}  \\ 
			MotionCtrl~\cite{wang2023motionctrl}     & \textcolor{goldorange}{\textbf{\checkmark}}      & \textcolor{goldorange}{\textbf{\checkmark}}         & \textcolor{graycolor}{\textbf{$\times$}}        \\
			CameraCtrl~\cite{he2024cameractrl}     & \textcolor{graycolor}{\textbf{$\times$}}      & \textcolor{goldorange}{\textbf{\checkmark}}         & \textcolor{graycolor}{\textbf{$\times$}}        \\ 
            DragNUWA~\cite{dragnuwa}     & \textcolor{goldorange}{\textbf{\checkmark}}      & \textcolor{graycolor}{\textbf{$\times$}}        & \textcolor{graycolor}{\textbf{$\times$}}        \\ 
            DragAnything~\cite{wu2024draganything}     & \textcolor{goldorange}{\textbf{\checkmark}}      & \textcolor{graycolor}{\textbf{$\times$}}         & \textcolor{graycolor}{\textbf{$\times$}}        \\ 
            Motion-I2V~\cite{shi2024motion}     & \textcolor{goldorange}{\textbf{\checkmark}}      & \textcolor{graycolor}{\textbf{$\times$}}        & \textcolor{goldorange}{\textbf{\checkmark}}       \\ 
            MOFA-Video~\cite{niu2024mofa}     & \textcolor{goldorange}{\textbf{\checkmark}}      & \textcolor{graycolor}{\textbf{$\times$}}        & \textcolor{graycolor}{\textbf{$\times$}}        \\
            TrackGo~\cite{zhou2024trackgo}      & \textcolor{goldorange}{\textbf{\checkmark}}      & \textcolor{graycolor}{\textbf{$\times$}}        & \textcolor{graycolor}{\textbf{$\times$}}        \\ \hline
            I2VControl (Ours)         & \textcolor{goldorange}{\textbf{\checkmark}}      & \textcolor{goldorange}{\textbf{\checkmark}}         & \textcolor{goldorange}{\textbf{\checkmark}}          \\ \hline
			\toprule
		\end{tabular}
	\end{center}
        \vspace{-6mm}
	\caption{Feature Comprehensiveness Comparison.}
        \vspace{-8mm}
	\label{comp:Comprehensiveness}
\end{table}

\subsection{Feature Comprehensiveness Analysis}
\label{sec:experiments:Comprehensiveness}

In this section, we theoretically compare the feature comprehensiveness between our method and some very recent baseline methods in Tab.~\ref{comp:Comprehensiveness}. 
For \textbf{motion drag}, users can select some points/objects on the input image and drag them to target positions, thus guide the video generation process. For \textbf{camera extrinsics} control, users are allowed to set a camera pose sequence to decide the camera movement of the generated video. For \textbf{motion brush}, we adopt the definition from motion-I2V~\cite{shi2024motion}, which enables the selected region to move (without specifying a trajectory or direction), while not affecting other unselected areas.
From Tab.~\ref{comp:Comprehensiveness}, we can see that only our method can deal with all the above common control situations in one single framework.

\subsection{Results and Comparisons}
\label{sec:experiments:comparisons}

In this section, we present our experimental results and comparisons. In Sec.~\ref{sec:experiments:creative}, we show the diverse capabilities of our framework. Then we compare with previous methods in Sec.~\ref{sec:experiments:comparison_works}. We also provide our results on another base model to demonstrate our adaptive nature in Sec.~\ref{sec:experiments:adaptive}.

\subsubsection{Diverse Capabilities and Creative Exploration}
\label{sec:experiments:creative}

As an disentangled and unified framework, I2VControl can combine multiple spatial control ideas without conflict, which can greatly meet the needs of video creation users. We show some examples in Fig.~\ref{fig:diversity} (please see the captions in the figure for textual explanation). The examples provided sufficiently demonstrate that our system empowers users with tools to unleash their creativity, encouraging them to explore and devise more intriguing control combinations.

\begin{figure}[t]
\centering
\vspace{-4mm}
\includegraphics[width=0.5\textwidth]{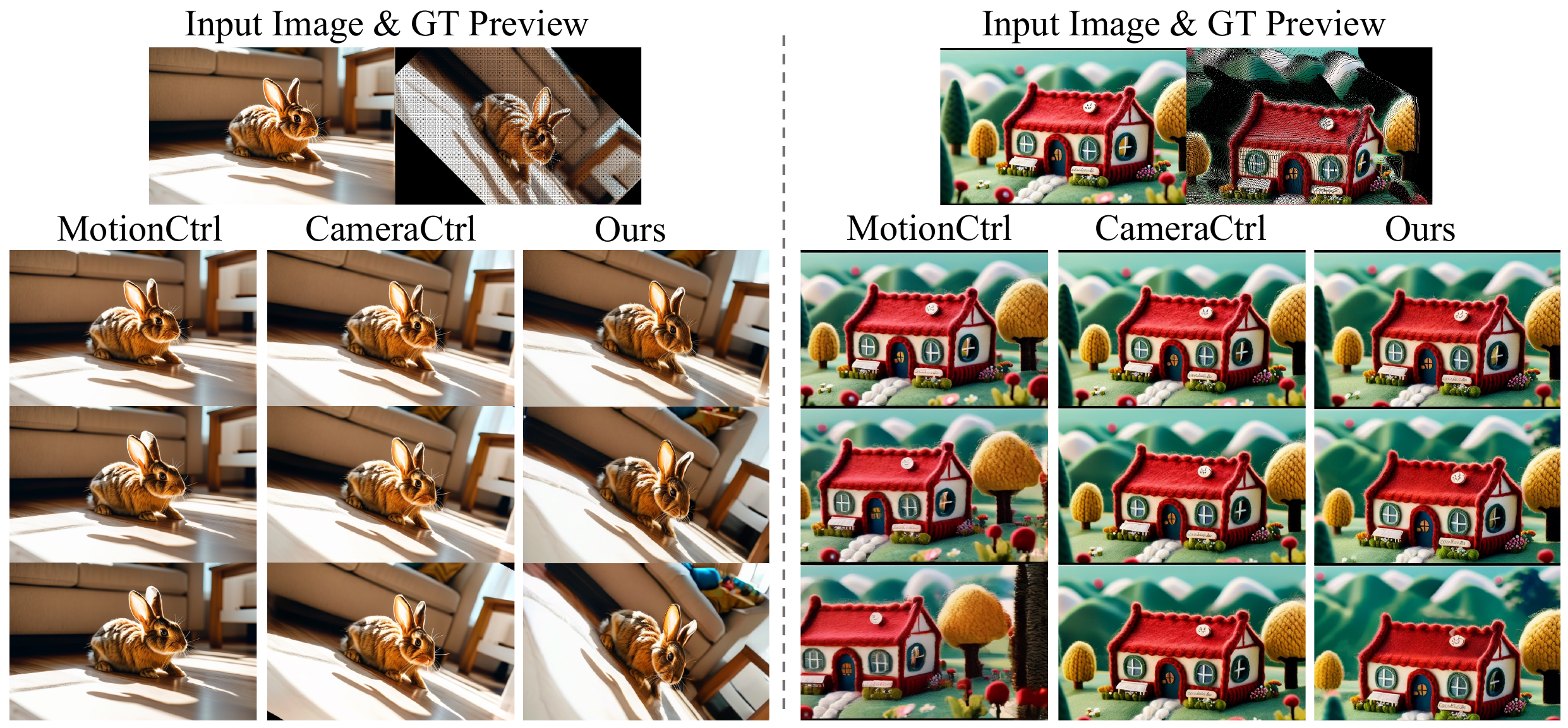}
\vspace{-4mm}
\caption{Comparison on camera control.}
\vspace{-6mm}
\label{fig:camera_comparison}
\end{figure}

\subsubsection{Comparison with Previous Works}
\label{sec:experiments:comparison_works}

We compare with recent methods including DragAnything~\cite{wu2024draganything}, MOFA-Video~\cite{niu2024mofa}, MotionCtrl~\cite{wang2023motionctrl}, CameraCtrl~\cite{he2024cameractrl}, and Motion-I2V~\cite{shi2024motion}. It is difficult to compare all methods under exactly the same experimental setup (the base models are inconsistent; even if the base models are aligned, the training data are not the same). We strive to reduce inconsistencies and train MotionCtrl and CameraCtrl on the same base model and data as ours. For the remaining methods, although the base models are different, they are at a similar level, all of which are U-Net methods (rather than recently developed high-quality DiT models). Furthermore, we will analyze the effect differences and point out that these differences are obviously not caused by the base model, but by the methods themselves.

Due to methodological limitations of the comparing methods, it is infeasible to compare them with ours under a unified control (like the $7$-th sample of Fig.~\ref{fig:diversity}), which is exactly the best proof of our superiority. Nevertheless, we still choose suitable datasets for them to enable meaningful (albeit suboptimal) comparisons with us.

\noindent \textbf{$\bullet$ VIPSeg and Manual Dragging Datasets}. 
We compare with very recent methods DragAnything~\cite{wu2024draganything} and MOFA-Video~\cite{niu2024mofa} on the VIPSeg~\cite{miao2022large} dataset and Manual Dragging (short as MD) dataset. For each sample, we divide the motion units on the initial frame. To construct the control signal for the comparing methods, we extract representative trajectories on each single units, where one trajectory is selected for the drag-units and four trajectories are selected for the borderland part. Quantitative results are in Tab.~\ref{tab:trajectory}, which shows that our method is superior to previous methods on both controlling precision and fidelity. The qualitative results can be seen in Fig.~\ref{fig:trajectory}. In the left sample, we use a pan-right camera movement and drag the object to the right, and only our result is right, thanks to our dense and disentangled nature. The sample on the right shows our good performance for rotation control.

\begin{table}[h]
\small
\setlength\tabcolsep{3.5pt}
\begin{center}
\begin{tabular}{c|c|ccc}
\bottomrule
\multirow{1}{*}{Dataset} &\multirow{1}{*}{Metric}   & \multirow{1}{*}{DragAnything}     & \multirow{1}{*}{MOFA-Video}   & \multirow{1}{*}{Ours}     \\ \hline
\multirow{3}{*}{VIPSeg}     & \texttt{ObjMC}$\downarrow$                  & 211.01   & 239.76 & \textbf{197.60}  \\  
                        & \texttt{FID}$\downarrow$                  & 136.56  & 141.50 & \textbf{127.44} \\ & \texttt{User}$\uparrow$                  & 1.92  & 2.63 & \textbf{3.87}
                        \\ \hline
\multirow{3}{*}{MD}    & \texttt{ObjMC}$\downarrow$                  & 31.24  & 43.60 & \textbf{17.07}    \\ 
                        & \texttt{FID}$\downarrow$                & 235.31 & 251.78 & \textbf{226.08} \\ 
                        & \texttt{User}$\uparrow$                & 2.45 & 2.93 & \textbf{4.21} \\
                        \toprule
\end{tabular}
\end{center}
\vspace{-4mm}
\caption{Comparison on VIPSeg and MD datasets.}
\vspace{-3mm}
\label{tab:trajectory}
\end{table}

\noindent \textbf{$\bullet$ RealEstate10K and MO Datasets}. 
We compare with MotionCtrl and CameraCtrl, on RealEstate10k (short as RE10K) and MO datasets in Tab.~\ref{camera}. Our quantitative results are the best, proving the precision and quality of our results. The qualitative results are shown in Fig.~\ref{fig:camera_comparison}, showing our good control precision again.

\begin{table}[th]
	\footnotesize
\begin{center}
\begin{tabular}{c|c|ccc}
\bottomrule
\multirow{1}{*}{Dataset} &\multirow{1}{*}{Metric}   & \multirow{1}{*}{MotionCtrl}     & \multirow{1}{*}{CameraCtrl}   & \multirow{1}{*}{Ours}     \\ \hline
\multirow{4}{*}{RE10K}     & \texttt{RotErr}$\downarrow$                  & 2.69   & 1.19 & \textbf{0.52}  \\  
                        & \texttt{TransErr}$\downarrow$                  & 12.31  & 22.72 & \textbf{9.93} \\ 
                        & \texttt{FID}$\downarrow$                  & 169.87   & 157.54 & \textbf{153.81}  \\  & \texttt{User}$\uparrow$                  & 2.13   & 2.01 & \textbf{4.31} 
                        \\ \hline
\multirow{4}{*}{MO}    & \texttt{RotErr}$\downarrow$                  & 2.09   & 1.58 & \textbf{1.11}  \\  
                        & \texttt{TransErr}$\downarrow$                  & 8.03  & 13.15 & \textbf{7.32} \\ 
                        & \texttt{FID}$\downarrow$                  & 96.13   & 97.36 & \textbf{92.91}  \\  & \texttt{User}$\uparrow$                  & 3.01   & 3.27 & \textbf{4.07} 
                        \\ 
                        \toprule
\end{tabular}
\end{center}
\vspace{-4mm}
\caption{Comparison on RealEstate10K and MO datasets.}
	\label{tab:time_size}
\label{camera}
\end{table}


\begin{figure}[t]
\centering
\includegraphics[width=0.5\textwidth]{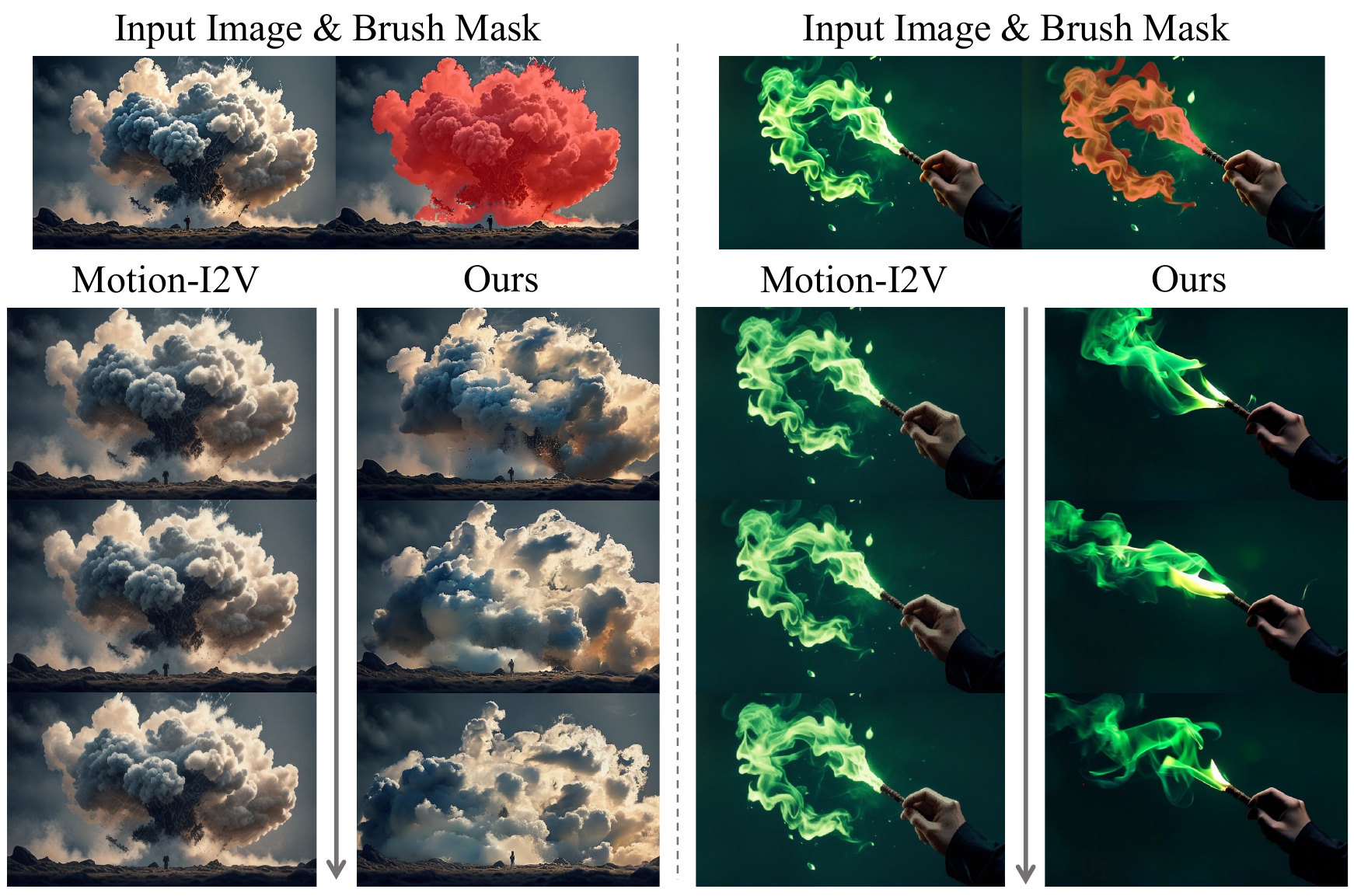}
\caption{Comparison on motion brush for fluid effects.}
\vspace{-2mm}
\label{fig:fluid_comparison}
\end{figure}

\begin{figure}[t]
\centering
\includegraphics[width=0.5\textwidth]{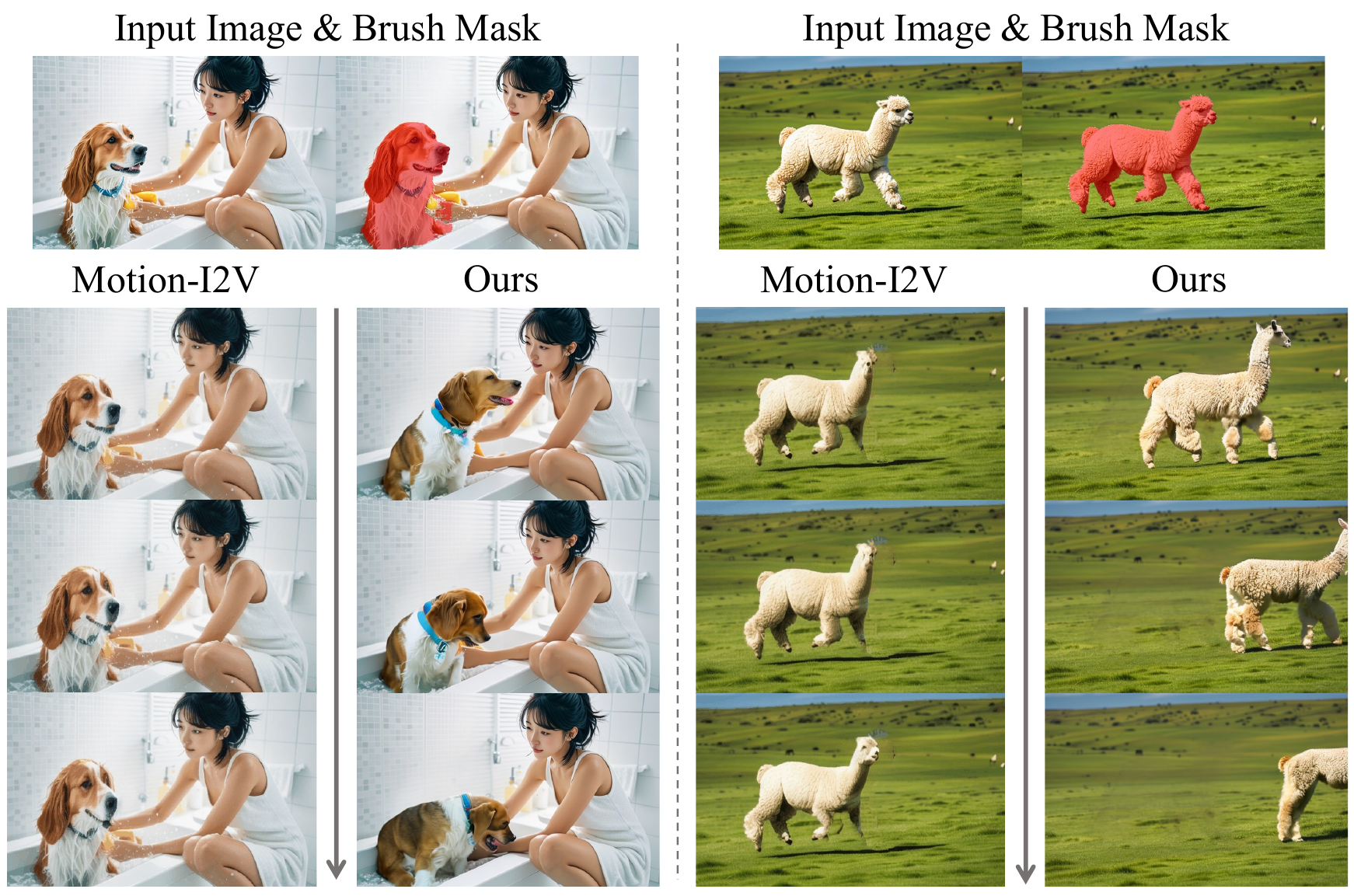}
\caption{Comparison on motion brush for object movements.}
\vspace{-2mm}
\label{fig:object_comparison}
\end{figure}

\noindent \textbf{$\bullet$ Object/Fluid Brush Datasets}.
We compare on the motion brush control task, where we brush masks on images and set scalar motion strength for the selected region without a motion trajectory or direction guide. We test on two datasets, the movable objects and fluid effects. We outperform previous method Motion-I2V on both datasets and all metrics (see Tab.~\ref{tab:brush}), reflecting our effectiveness better response to masks and more significant motion. The numerical difference is very intuitively reflected in the visualization results, see Fig.~\ref{fig:object_comparison} and Fig.~\ref{fig:fluid_comparison}. The result of Motion-I2V has a small motion range, and the movable pixels seems not able to move out of the given mask. This might be related to the control stragety based on optical flow. In contrast, our method can produce very large-scale motion. For example, in Fig.~\ref{fig:fluid_comparison}, the flame emitted from the wand can be transformed into different shapes. The brushed objects can also move away from the origin mask, such as the animal on the right of Fig.~\ref{fig:object_comparison}.

\begin{table}[h]
\small
\setlength\tabcolsep{5pt}
\begin{center}
\begin{tabular}{c|c|cc}
\bottomrule
\multirow{1}{*}{Dataset} &\multirow{1}{*}{Metric}   & \multirow{1}{*}{Motion-I2V}      & \multirow{1}{*}{Ours}     \\ \hline
\multirow{4}{*}{Object Brush}     & \texttt{MSC}$\uparrow$                  & 6.87   & \textbf{64.59}  \\  
                        & \texttt{IoU($\%$)}$\uparrow$                  & 15.43  & \textbf{60.34} \\ 
                        & \texttt{FID}$\downarrow$                  & 246.92   & \textbf{207.68}  \\  & \texttt{User}$\uparrow$                  & 0.34   & \textbf{4.12} 
                        \\ \hline
\multirow{4}{*}{Fluid Brush}    & \texttt{MSC}$\uparrow$                  & 2.57   & \textbf{97.29}  \\  
                        & \texttt{IoU($\%$)}$\uparrow$                  & 5.60  & \textbf{46.37} \\ 
                        & \texttt{FID}$\downarrow$                  & 266.39   & \textbf{223.22}  \\  & \texttt{User}$\uparrow$                  & 1.07   & \textbf{4.56} 
                        \\ 
                        \toprule
\end{tabular}
\end{center}
\vspace{-4mm}
\caption{Comparison on Object/Fluid Brush datasets.}
\label{tab:brush}
\end{table}

\subsubsection{Adaptive Nature}
\label{sec:experiments:adaptive}

To show our adaptive nature, we also train on other base models. Specifically, we train on Seedance~\cite{gao2025seedance}(see visual results \href{https://wanquanf.github.io/I2VControl}{here}) and Wanx 2.1~\cite{wan2.1}(we have released our code base on Wan 2.1). Please note that the implementation of attention varies depending on the specific base model used. However, the organization of the control signals and the encoder remain consistent. Our structure can still work well on these base models, which proves our adaptive nature.

%% file: 06_conclusion.tex
\section{Conclusion}
\label{sec:conclusion}

In this work, we explore the motion controllability of image-to-video synthesis, a vital aspect for enhancing end-user utility and solving logic conflicts in previous methods. Our proposed framework, \textbf{\papername}, implements flexible video motion controllers with disentangled control signals, achieving a user-friendly interface and outstanding performance in various motion control tasks including motion brush, object dragging, and camera pose controlling. For the future, we aim to enhance the framework to handle increasingly complex scenarios, integrating more nuanced control over video motion not only at the structural but also at the textual level. Additionally, we also expect ongoing advancements in model architectures to boost the capabilities and operational efficiency of our framework.

\small \noindent{\bf Acknowledgement} \textcolor{black}{We thank Yuxi Xiao and Yudong Guo for their generous help and advice with the tracking algorithm. We also thank Hongrui Cai for developing the open-source version of the model and training code. Finally, we thank Jianzhu Guo for insightful suggestions on the gradio demo.} 